\documentclass[sigconf,screen]{acmart}

\AtBeginDocument{%
  \providecommand\BibTeX{{%
    \normalfont B\kern-0.5em{\scshape i\kern-0.25em b}\kern-0.8em\TeX}}}

\usepackage{amsmath}
\usepackage{graphicx}
\usepackage{array}
\usepackage{booktabs}

\usepackage{algorithm}

\usepackage{amsfonts}
\usepackage{enumitem}
\usepackage{hyperref}
\usepackage{xcolor}
\newcolumntype{P}[1]{>{\centering\arraybackslash}p{#1}}

\usepackage{subfigure}

\usepackage{natbib}
\usepackage{float}
\usepackage{stfloats}

\newcommand{\sets}[1]{\ensuremath{\mathcal{#1}}}

\theoremstyle{definition}
\newtheorem{definition}{Definition}[section]

\copyrightyear{2023}
\acmYear{2023}
\setcopyright{acmlicensed}\acmConference[AIES '23]{AAAI/ACM Conference on AI, Ethics, and Society}{August 8--10, 2023}{Montréal, QC, Canada}
\acmBooktitle{AAAI/ACM Conference on AI, Ethics, and Society (AIES '23), August 8--10, 2023, Montréal, QC, Canada}
\acmPrice{15.00}
\acmDOI{10.1145/3600211.3604664}
\acmISBN{979-8-4007-0231-0/23/08}

\allowdisplaybreaks

\begin{document}

\title{Learning Optimal Fair Decision Trees: Trade-offs Between Interpretability, Fairness, and Accuracy}


\author{Nathanael Jo}
\email{nathanael.jo@gmail.com}
\affiliation{%
  \institution{USC Center for AI in Society}
  \city{Los Angeles, CA}
  \country{USA}}

\author{Sina Aghaei}
\email{saghaei@usc.edu}
\affiliation{%
  \institution{USC Center for AI in Society}
  \streetaddress{3715 McClintock Ave GER 240}
  \city{Los Angeles}
  \state{CA}
  \country{USA}
  \postcode{90089}}

\author{Jack Benson}
\email{jackhbenson17@gmail.com}
\affiliation{%
  \institution{USC Center for AI in Society}
  \streetaddress{3715 McClintock Ave GER 240}
  \city{Los Angeles}
  \state{CA}
  \country{USA}
  \postcode{90089}}
  
\author{Andr\'es G\'omez}
\email{andgomez@usc.edu}
\affiliation{%
  \institution{University of Southern California}
  \streetaddress{3715 McClintock Ave GER 240}
  \city{Los Angeles}
  \state{CA}
  \country{USA}
  \postcode{90089}}

\author{Phebe Vayanos}
\email{phebe.vayanos@usc.edu}
\affiliation{%
  \institution{USC Center for AI in Society}
  \streetaddress{3715 McClintock Ave GER 240}
  \city{Los Angeles}
  \state{CA}
  \country{USA}
  \postcode{90089}}

\renewcommand{\shortauthors}{Jo, et al.}

\begin{abstract}
  The increasing use of machine learning in high-stakes domains -- where people's livelihoods are impacted -- creates an urgent need for interpretable, fair, and highly accurate algorithms. With these needs in mind, we propose a mixed integer optimization (MIO) framework for learning optimal classification trees -- one of the most interpretable models -- that can be augmented with arbitrary fairness constraints. In order to better quantify the ``price of interpretability'', we also propose a new measure of model interpretability called \textit{decision complexity} that allows for comparisons across different classes of machine learning models. We benchmark our method against state-of-the-art approaches for fair classification on popular datasets; in doing so, we conduct one of the first comprehensive analyses of the trade-offs between interpretability, fairness, and predictive accuracy. Given a fixed disparity threshold, our method has a price of interpretability of about 4.2 percentage points in terms of out-of-sample accuracy compared to the best performing, complex models. However, our method consistently finds decisions with almost full parity, while other methods rarely do. 
\end{abstract}





\maketitle

\section{Introduction}
There is growing interest in using machine learning (ML) to make decisions in high-stakes domains. For instance, ML algorithms are now commonly used to determine a criminal's risk of recidivism in the United States~\citep{angwin_larson_mattu_kirchner_2016}. There is also a growing literature in designing algorithms to determine the best course of action for homeless individuals \citep{azizi2018designing}, to diagnose and treat of various illnesses \cite{fatima2017survey}, and many more. In these contexts, it is necessary for such models to be both \textit{accurate} (in order to minimize erroneous predictions that negatively affect stakeholders) and \textit{interpetable} (so that decisions are transparent and hence accountable). We therefore focus our attention to the problem of learning optimal classification trees. Classification trees are among the most interpretable of models \cite{rudin2019stop}, and optimal trees -- rather than ones that are built using heuristics -- maximize predictive accuracy. They belong to a broader set of models known as decision trees (the other type of decision tree being regression trees, which apply to datasets with real valued labels).

There exist numerous qualitative ways of characterizing interpretability \citep{murdoch2019definitions}. These notions often involve querying humans (especially practitioners, community stakeholders, etc.), so that interpretability desiderata can be tailored to the application or population at hand \citep{doshi2017towards}. Unfortunately, quantitative notions of interpretability are lacking in the machine learning literature, making it hard to compare models systematically without humans in the loop. One such measure -- sparsity -- is one of the only quantitative proxies for interpretability, but does not allow for equivalent comparisons between different model classes \citep{rudin2022interpretable} (see Section~\ref{section:related_works} for more details). Therefore, we seek to address this gap by proposing  a new notion of interpretability in order to more formally quantify the price of interpretability between our classification trees and more complex models.

Apart from interpretability, another crucial consideration in machine learning for high-impact situations is \textit{fairness}. After all, an algorithm that affects people's well-being should be aware of the particular historical and/or social contexts that surround the learning problem. However, what constitutes as fair may vary widely depending on one's goals, domains of interest, etc. This discussion surrounding ``algorithmic fairness'' consists of a rich literature that bridges philosophy, economics, and computer science \citep{rawls2004theory, dwork2012fairness, roemer2015equality}. Consider, for example, a na\"ive approach where a classifier learns from data where the sensitive attributes are omitted (such as race, sex, and others). Also known as ``fairness through unawareness'', this approach often leads to discrimination because some other non-sensitive attribute(s) may be correlated with the withheld features (e.g., race is inextricably linked to ZIP code and income) \citep{pedreshi2008discrimination}.

Another approach -- and one that this paper adopts -- is the notion of ``group'' statistical fairness. A classifier satisfying group fairness is now aware of the sensitive features in the data but will only produce decisions that enforce parity between segments of the population. This is in contrast to ``individual'' fairness, which requires that individuals with similar characteristics be classified similarly \citep{dwork2012fairness}. Both perspectives have their advantages and drawbacks, but in this work we will focus on group fairness primarily because of it is simpler to define and easier for stakeholders to understand; as such, many practitioners in practice value assessing and enforcing group fairness (see, e.g.,~\cite{lahsa2018report}). In the following sections, we will discuss in more depth the use of many notions of group fairness in the machine learning literature.

\subsection{Problem Statement}\label{section:problem_statement}
We now formalize the problem we study. Let $\sets D:= \{(x^i, y^i)\}_{i \in \mathcal{I}}$ be training data indexed in the set $\sets{I} \in \{1, \dots, I\}$. Each datapoint $i \in \sets{I}$ consists of a vector $x^i$ with $F$ features (i.e., $x^i \in \mathbb{R}^F$), and a class label $y^i \in \sets{K}$, where $\sets{K}$ is the finite set of possible classes. Throughout this paper, we will consider the case of binary classes, i.e., $\sets K:=\{0,1\}$ where $y=1$ will be referred to as the positive class. Note that the literature on fairness for multi-class learning is limited, see e.g., \citet{denis2021fairness}. The goal is to learn, over all possible trees of maximum depth $d$, the tree that maps $x^i$ to $y^i$ and maximizes out-of-sample accuracy, using in-sample performance as a proxy. Further, suppose the population can be divided into different sensitive groups (whereby discrimination exists or is a concern within the classification problem, e.g., race, gender). Let $\sets{P}$ denote levels of a sensitive attribute, and let each datapoint $i$ have a value $p^i \in \sets{P}$. The features in $\sets{P}$ may or may not be included in the vector $x^i$ since there may be legal or ethical considerations barring the use of protected features in the predictive task \citep{chen2019fairness}.

\subsection{Common Notions of Group Fairness}\label{section:fairness_def}
In this section, we define five common notions of group fairness in the machine learning literature that we will use. Let $\hat{Y}$ be the classifier's prediction, and let $X$, $Y$, $\hat{Y}$, and $P$ be random variables for features, classes, classifier's predictions, and protected features, respectively; their joint distribution is unknown and denoted by $\mathbb{P}$.

\noindent \textbf{Statistical Parity.}
A classifier satisfies statistical parity if the probability of receiving a positive class is equal across all protected groups \citep{dwork2012fairness}. Formally, this means
\begin{equation*}
\mathbb{P}[\hat{Y} = 1 | P=p] = \mathbb{P}[\hat{Y} = 1 | P=p'] \quad \forall p, p' \in \sets{P}.
\end{equation*}

\noindent \textbf{Conditional Statistical Parity.} A classifier satisfies \textit{conditional} statistical parity if the probability of receiving a positive class is equal across all protected groups, conditional on some legitimate feature(s) indicative of risk \citep{corbett2017algorithmic}. This may be considered as a fairer notion than statistical parity because it takes into account the distribution of risk factors within each sensitive group. Letting $L$ (which is a subvector of $X$) represent the random variable taken from a set $\sets{L}$ of legitimate features, conditional statistical parity is satisfied if 
\begin{equation*}
\begin{split}
\mathbb{P}[\hat{Y} = 1 | P=p, L=\ell] = \mathbb{P}[\hat{Y} = 1 | P=p', L=\ell] \\ \quad \forall p, p' \in \sets{P}, \ell \in \sets{L}.
\end{split}
\end{equation*}

\noindent \textbf{Predictive Equality.} A classifier satisfies predictive equality if all protected groups have the same false positive rates (FPR) \citep{chouldechova2017fair}, i.e., 
\begin{equation*}
\mathbb{P}[\hat{Y} = 1 | P=p, Y=0] = \mathbb{P}[\hat{Y} = 1 | P=p', Y=0] \quad \forall p, p' \in \sets{P}.
\end{equation*}

\noindent \textbf{Equal Opportunity.} A classifier satisfies equal opportunity if all protected groups have the same true positive rate (TPR) \citep{hardt2016equality}. The formal definition of equal opportunity is 
\begin{equation*}
\mathbb{P}[\hat{Y} = 1 | P=p, Y=1] = \mathbb{P}[\hat{Y} = 1 | P=p', Y=1] \quad \forall p, p' \in \sets{P}.
\end{equation*}

\noindent \textbf{Equalized Odds.} Equalized odds combines predictive equality and equal opportunity such that both FPR and TPR must be similar across all protected groups \citep{hardt2016equality}. Equalized odds is, of course, a stronger condition than predictive equality and equal opportunity individually. Formally, equalized odds is satisfied if \begin{equation*}
\begin{split}
\mathbb{P}[\hat{Y} = 1 | P=p, Y=y] = \mathbb{P}[\hat{Y} = 1 | P=p', Y=y] \\ \forall p, p' \in \sets{P}, y \in \{0, 1\}.
\end{split}
\end{equation*}

\subsection{Related Works}\label{section:related_works}
Our work relates to five streams of literature in machine learning, which we review in turn. In this section we briefly review related works in the machine learning literature. 

\textbf{Discrimination Prevention in Machine Learning.} The literature on fairness in machine learning is extensive -- we point the interested reader to \cite{mehrabi2021survey} and \cite{vzliobaite2017measuring} for surveys of this topic. In general, there are three approaches that existing works employ in order to promote fairness. The first is \textit{pre-processing}, which entails eliminating discrimination within the training data, see \cite{zemel2013learning, feldman2015certifying, hajian2012methodology, kamiran2013quantifying}, among others. The second is \textit{post-processing}, which takes the model's output and alters its decisions in ways that promote some fairness metric, see \cite{kamiran2010discrimination, hardt2016equality, pleiss2017fairness}. Finally, one may also employ \textit{in-processing} by modifying existing models so that fairness is integrated within its learning goal; for instance, approaches have utilized regularization \cite{berk2017convex, kamishima2012fairness, aghaei2019learning} or different heuristics \cite{kamiran2010discrimination, calders2010three}. Our work falls under the category of in-processing discrimination prevention techniques.

\textbf{Fair Decision Trees.} Within the stream of literature on discrimination prevention, our work most closely relates to approaches for learning fair decision trees. For instance, Ranzato et al.~\cite{ranzato2021fair} adapt a genetic algorithm for learning robust decision trees in order to prioritize individual fairness. In the online setting, Zhang et al.~\cite{zhang2019faht} propose a fairness-aware Hoeffding tree that builds decision trees over streams of data. Grari et al.~\cite{grari2019fair} take an adversarial approach to training fair gradient-boosted trees that promote statistical parity. Kanamori and Arimura~\cite{hi2019fairness} propose a post-processing step that uses mixed-integer optimization (MIO) to edit the branching thresholds of the tree's internal nodes to satisfy some fairness constraint; in the paper they use statistical parity and equalized odds. In a similar manner, Zhang et al.~\cite{zhang2020fairness} flip the outcomes of different paths of a learned tree in order to improve fairness. Closely related to our work is that of Aghaei et al.~\cite{aghaei2019learning}, who propose a mixed-integer optimization framework to build optimal and fair decision trees that prevent disparate impact and/or disparate treatment. In contrast to our work, Aghaei et al.~\cite{aghaei2019learning} enforces fairness via regularization while our method uses constraints in the optimization. Finally, Kamiran et al.~\cite{kamiran2010discrimination} propose a two-pronged approach: the first is to incorporate sensitivity gain (IGS) with information gain (IGC) as a heuristic to build a fairer decision tree, while the second is to relabel the predictions in the tree's leaf nodes to further promote statistical parity.

\textbf{Optimal Decision Trees.} Mixed-integer optimization (MIO) has recently gained traction as a framework to solve various machine learning problems. Particularly related to this paper are approaches that use MIO to learn optimal decision trees in order to improve on the traditional classification/regression tree (CART) algorithms \cite{breiman2017classification}, which rely on a heuristic. Many works introduce novel formulations for learning optimal classification trees \cite{aghaei2021strong, aghaei2019learning, verwer2019learning, bertsimas2017optimal}. Elmachtoub et al.~\cite{elmachtoub2020decision} use MIO to learn decision trees that minimize a loss function derived from the ``predict-then-optimize'' framework. There exist several extensions to learning decision trees using MIO as well. For instance, Mi{\v{s}}i{\'c}~\cite{mivsic2020optimization} uses MIO to solve tree ensemble models to optimality; Jo et al.~\cite{jo2021learning} to learn optimal prescriptive trees from observational data; and Justin et al.~\cite{justin2021optimal} and Bertsimas et al.~\cite{bertsimas2019robust} to learn optimal robust decision trees. In this paper, we build on the MIO method introduced by Aghaei et al. \cite{aghaei2021strong} by conducting extensive experiments when we add various fairness constraints to the model.

\textbf{Notions of Interpretability.} There exist numerous proxies or desiderata for interpretability in the machine learning literature, including:
\begin{itemize}
    \item \textbf{Sparsity}: The simplicity of a model. There are many ways to define sparsity, which also differ across model classes. For instance, within the context of decision trees, Rudin et al.~\citet{rudin2022interpretable} define sparsity using the number of leaves, where trees with fewer leaves are sparser and thus preferable. In a regression model, sparsity is widely associated with the number of nonzero regression coefficients \citep{tibshirani1996regression,akaike1974new,miller2002subset}. In general, a numeric value for sparsity is only useful when comparing models within the same class. This is because some model classes have sparsity that does not grow uncontrollably (e.g., regression is constrained by the number of features), while others are a function of design and data (e.g., trees become increasingly complex as depth increases) \citep{rudin2022interpretable}.
    \item \textbf{Simulatability}: The extent to which a human can internally simulate and reason about part of the entire decision-making schema \citep{murdoch2019definitions}. Shallow decision trees are some of the most simulatable models since we can easily visualize and understand if-else rules. This is in contrast to a neural network, where the numerous connections between nodes result in complicated calculations that a human cannot keep track of.
    \item \textbf{Scope} (Global vs.\ Local interpretability): Global interpretability means that a human can wholly understand the decision schema (as is the case with, say, regression, where everything one needs to know about the model is encoded in its coefficients). On the other hand, local interpretability means that one could reason about how and why a particular datapoint gets classified a certain way \citep{molnar2022}. For instance, while the entire behavior of a k-nearest neighbor (kNN) algorithm is incomprehensible, especially when a dataset is large, humans can reason about local behavior: a datapoint is classified a certain way because most of its k-nearest neighbors are classified the same way.
\end{itemize}

There are many other proxies (e.g., uncertainty, algorithmic transparency, monotonicity, etc.), as well as other notions of interpretabilty that apply to various stages of the predictive pipeline (e.g., considerations in feature engineering). We refer the interested reader to \cite{murdoch2019definitions,rudin2022interpretable,molnar2022} for an extensive overview of interpretable machine learning. As we have mentioned previously, interpretability is broad, loosely-defined, and often context dependent. Nonetheless, existing quantitative measures (such as sparsity) are only useful for comparisons within the same model class; our work attempts to address this shortcoming by proposing a new measure of interpretability that allows for comparisons \textit{across} model classes.

\textbf{Interpretability, Accuracy, and Fairness.} Several works have explored the possible trade-offs between predictive accuracy and interpretability (without considering fairness), often in application-specific contexts~\citep{johansson2011trade, baryannis2019predicting} or within a certain model class~\citep{akaike1974new,schwarz1978estimating, alcala2006hybrid, you2022interpretability} where interpretability desiderata differ widely. In a more general setting, \citet{dziugaite2020enforcing} propose a learning framework via empirical risk minimization that imposes interpretability constraints, and characterizes when trade-offs between accuracy and interpretability may exist -- this is in line with other works that find the optimal model within a specified model class subject to interpretability constraints (e.g., \citet{azizi2018designing, rudin2022interpretable}). On the other hand, \citet{semenova2022existence} allow for comparisons \textit{across} various model classes by using Rashomon sets to gauge the likelihood of simpler models with competing accuracy performance existing within a hypothesis space. However, none of these works consider fairness as a dimension in model selection. There is a dearth of literature touching the trade-offs between all three: interpretability, accuracy, \textit{and} fairness. \citet{agarwal2021trade} finds theoretical results that in general, there exist more complex models that perform strictly better with respect to fairness and accuracy. Our work extends this finding by experimentally characterizing these trade-offs. \citet{wang2022pursuit} compare models with varying levels of interpretability in predicting criminal recidivism, considering notions of fairness where the prediction task is continuous (e.g., calibration). In contrast, our work considers binary fairness notions and we run experiments more generally on various benchmark datasets -- which include predicting recidivism. We additionally compare several fairness-promoting algorithms in the literature, while \citet{wang2022pursuit} only consider off-the-shelf ML models.

\subsection{Contributions}
In this work, we build upon an MIO formulation to learn optimal decision trees initially proposed in \citet{aghaei2021strong} by showcasing its flexible modeling power in considering various notions of fairness. We benchmark these experiments against two state-of-the-art methods that similarly learn fair decision trees, as well as three fair classification algorithms that are compatible with a variety of ML models. In order to more formally juxtapose the interpretability of ML models, we propose a new measure called \textit{decision complexity}. In doing so, we conduct one of the first experiments to characterize the trade-offs between performance and interpretability within the algorithmic fairness literature, and discuss the practical considerations among these dimensions. In particular, we observe that the best performing, complex models have on average 4.2 percentage points higher out-of-sample accuracy than our interpretable approach, indicating the price of interpretability. We also observe that our method is particularly well-suited in finding decisions with full parity, while other methods do not boast the same guarantees.

The remainder of this paper is structured as follows. In Section~\ref{sec:decision_complexity}, we introduce a new notion of interpretability -- decision complexity -- that allows for comparisons across model classes. We then outline the MIO formulation to learn optimal and fair decision trees in Section~\ref{section:formulation}. Finally, we conduct and analyze our computational experiments in Section~\ref{section:experiments}.

\section{Decision Complexity}\label{sec:decision_complexity}

As discussed in Section~\ref{section:related_works}, sparsity/simplicity is one of the only metrics in the literature to quantify interpretability, and is mainly relevant when comparing the models within the same class. In our work, however, we are concerned with comparing a measure of interpretability \textit{across} model classes (e.g., comparing decision trees with kNN). To the best of our knowledge, there is no such universal definition or framework for interpretability. We therefore propose a new notion that can quantify the interpretability of predictive models belonging either to the same or to different model classes.


\begin{definition}[Decision Complexity]
Given a trained classifier, decision complexity captures the minimum number of parameters needed for the classifier to make a prediction on a new datapoint.
\end{definition}

\begin{example}[Binary Classification Trees]
The decision complexity of binary classification trees is measured by the number of nodes in the tree (branching plus leaves), which corresponds to the number of times a datapoint is routed through the tree and how it is classified.
\end{example}

\begin{example}[Random Forest]
Building from binary classification trees, a random forest's decision complexity is equal to the sum of nodes (branching and leaves) over all trees in the forest.
\end{example}

\begin{example}[Linear/Logistic Regression]
Assuming full linearity (i.e., no interaction or quadratic terms), the decision complexity of simple regression models is always equal to the number of features in the data (plus a possible bias term).
\end{example}

\begin{example}[k-Nearest Neighbors]
In order to classify a new datapoint, we must find the distance between said datapoint with all training points, and therefore a kNN's decision complexity is equal to the size of the training data. While there exist more efficient algorithms in practice that do not require finding all pairwise comparisons, we are concerned primarily with how a human can walk through the algorithm's decisions rather than the computational complexity to train the classifier.
\end{example}

\begin{example}[Support Vector Machines]
Decision complexity in SVMs highly depends on the choice of kernel. Linear kernels have a complexity equal to the number of features in the data, which correspond to the coefficients in the hyperplane that separates classes. However, gaussian RBF kernels have a decision complexity equal to the number of support vectors in the training set. This is because, upon training, each of the support vectors are associated with weights that determine how a new datapoint gets classified.
\end{example}

\begin{example}[Neural Network]
A neural network's decision complexity is equal to the number of ``connections'' between all nodes, since a new datapoint is classified through matrix calculations using solved weights from each connection.
\end{example}

Decision complexity can be viewed as an extension to sparsity (in that sparsity is often used analogously to simplicity) but has the advantage of being general enough for all model classes. 
It can also be interpreted as attempting to quantify a part of simulatability -- the more decisions a model takes, the harder it tends to be for humans to simulate the decision making process. 

\section{Formulation}\label{section:formulation}

In this section, we present the MIO formulation to learn optimal decision trees proposed in \citet{aghaei2021strong}. The paper introduced the formulation without an emphasis on fairness, but in this work we use the formulation as a building block to which we add various fairness constraints. From hereon, we will refer to the combination of the MIO formulation and the fairness constraints as \textit{FairOCT}.

\subsection{From Decision Tree to Flow Graph}
\label{section:flow_graph}

\begin{figure}[ht]
\includegraphics[width=0.45\textwidth]{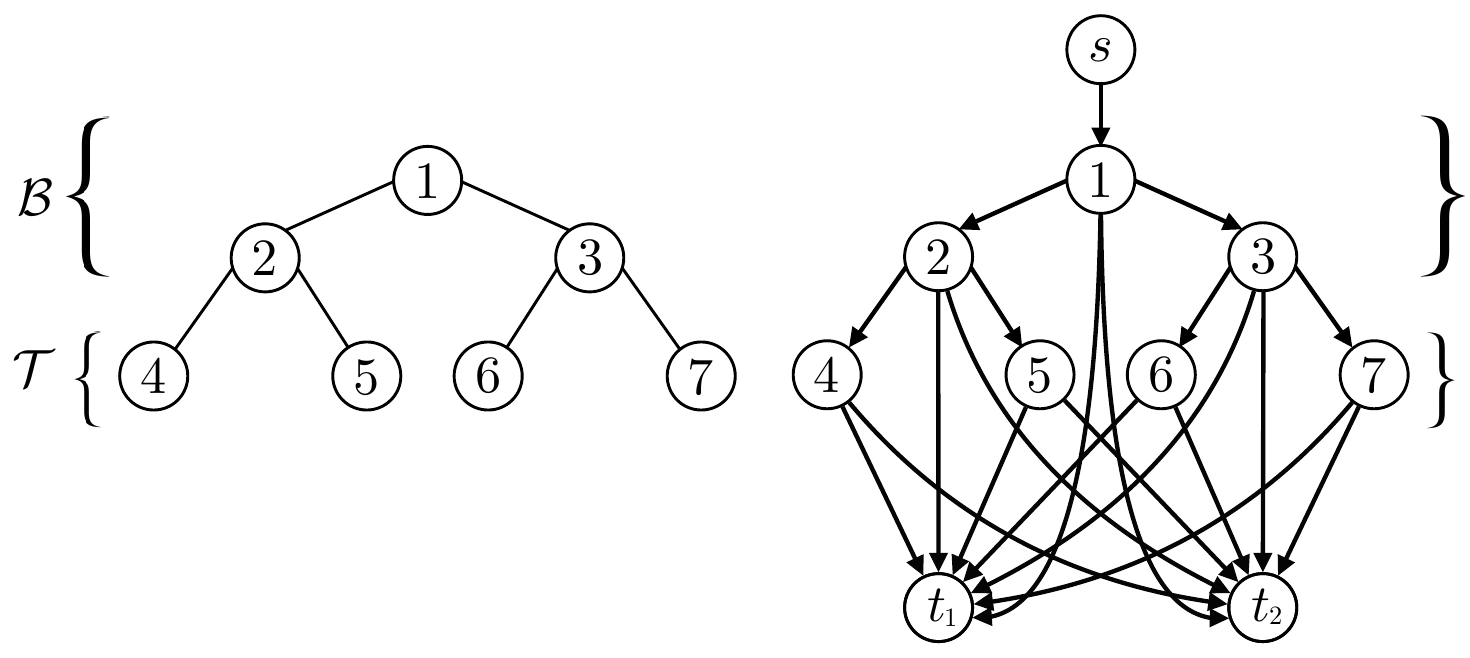}
\centering
\caption{A decision tree of depth 2 (left) and its associated flow graph (right) with two classes.}
\label{fig:flow_graph}
\end{figure}

We first introduce the modeling framework we use for decision trees. The key idea is to convert a decision tree into a directed, acyclic graph where all arcs ``flow'' from the tree's root to its leaves. We start with a perfect binary tree of depth~$d$, whose nodes are labeled $1$ through $(2^{d+1}-1)$ in order of a breadth-first search. Let $\sets{B} := \{1, \dots, 2^{d}-1\}$ denote the set of branching nodes and $\sets{T} := \{2^d, \dots, 2^{d+1}-1\}$ the set of terminal nodes. We then convert all arcs in the tree to point from the parent node to its child node (see Figure~\ref{fig:flow_graph}, left). From the binary tree, we connect a source $s$ to the root and all nodes $n \in \sets{B} \cup \sets{T}$ to sinks $t_k$, one for every class $k \in \sets{K}$ (see Figure~\ref{fig:flow_graph}, right). While we assume that we have binary classes, this formulation generalizes to arbitrary finite classes. All arcs have a capacity of 1, so each datapoint is weighted equally and flows from the source $s$ to one sink $t_k$, where $k$ is the class that the decision tree assigns to that datapoint. From hereon, we refer to this structure as a ``flow graph''.

\subsection{MIO Formulation}\label{section:mio}
With the flow graph at hand, we present the MIO formulation for learning optimal classification trees. Without loss of generality, we assume that features are binary, i.e., $x^i \in \{0,1\}^F$ -- this assumption can easily be relaxed to cater for integer or categorical features, see Remark~1 in~\cite{aghaei2021strong}. We encode the branching structure of the tree with variables $b_{nf} \in \{0, 1\}$, for all $n \in \sets{B}, f \in \sets{F}$, which indicate if feature~$f$ is selected for branching at node $n$. We also encode the prediction scheme of the tree using variables $w_{nk} \in \{0, 1\}$, for all $n \in \sets{B} \cup \sets{T}, k \in \sets{K}$, which equal 1 if and only if (iff) node $n$ assigns class $k$ to all datapoints that land on that node. We let $p_n \in \{0, 1\}$ indicate if node~$n$ is a prediction node, in which case it must assign a class to all datapoints that land on that node and no further branching is allowed. We also define ``flow variables''~$z$ to capture the flow of datapoints, where $z^i_{a(n), n} \in \{0, 1\}$ equals 1 iff datapoint $i$ flows from the direct ancestor of node $n$,~$a(n)$, to~$n$.

The formulation is as follows:

\begin{subequations}
\begin{flalign}
& \text{maximize} \;\;  \textstyle \sum_{i \in \mathcal I} \textstyle \sum_{n \in \sets B \cup \sets T } z^i_{n,t_{y^i}}  & \text{subject to}
\label{eq:flow_reg_all_obj}\\
& {\textstyle \sum}_{f \in \sets F}b_{nf} + p_n + \textstyle \sum_{m \in \sets A(n)}p_m = 1   &\hspace{-5cm}  \forall n \in \sets B \label{eq:flow_reg_all_branch_or_predict}\\
&  p_n+\textstyle \sum_{m \in \sets A(n)}p_{m} =1   &  \forall n \in \sets T \label{eq:flow_reg_all_terminal_leaf}\\
& \displaystyle z^i_{a(n),n} =  z^i_{n,\ell(n)} + z^i_{n,r(n)} + \sum_{k \in \sets K}z^i_{n,t_k}   &\hspace{-5cm}  \forall n \in \sets B, i \in \mathcal I \label{eq:flow_reg_all_conservation_internal}\\
&  \displaystyle z^i_{a(n),n} = \textstyle \sum_{k \in \sets K}z^i_{n,t_k}  &\hspace{-5cm}   \forall i \in \mathcal I,n \in \sets T \label{eq:flow_reg_all_conservation_terminal}\\
& \displaystyle z^i_{s,1} \leq 1 &\hspace{-5cm} \forall i \in \mathcal I\label{eq:flow_reg_all_source}\\
&  \displaystyle z^i_{n,\ell(n)}\leq \textstyle \sum_{f \in \sets F: x_{f}^i=0}b_{nf} &\hspace{-5cm} \forall n \in \sets B, i \in \mathcal I \label{eq:flow_reg_all_branch_left}\\
&  \displaystyle z^i_{n,r(n)}\leq \textstyle \sum_{f \in \sets F: x_{f}^i=1}b_{nf}  &\hspace{-5cm} \forall n \in \sets B, i \in \mathcal I \label{eq:flow_reg_all_branch_right}\\
&  \displaystyle z^i_{n,t_{k}} \leq  w^n_{k} &\hspace{-5cm} \forall i \in \mathcal I, n \in \sets B \cup \sets T,k \in \sets K \label{eq:flow_reg_all_sink}\\
&  \textstyle \sum_{k \in \sets K}w^n_{k} = p_n  &\hspace{-5cm}  \forall n \in \sets B \cup \sets T \label{eq:flow_reg_all_leaf_prediction}\\
&  \displaystyle w^n_{k} \in \{0,1\}  &\hspace{-5cm}   \forall n \in \sets B \cup \sets T,k \in \sets K \label{eq:integrality_w} \\
&  \displaystyle b_{nf} \in \{0,1\}  &\hspace{-5cm}   \forall n \in \sets B,f \in \sets F \\
&  \displaystyle p_{n} \in \{0,1\}  &\hspace{-5cm}   \forall n \in \sets B \cup \sets T \\
&  \displaystyle z^i_{a(n),n}, z^i_{n,t_k}\in \{0,1\}  &\hspace{-5cm}  \forall n \in \sets B \cup \sets T,i \in \sets I, k \in \sets K,
\end{flalign}
\label{eq:flow_reg_all}
\end{subequations}
where $\ell(n)$ (resp.\ $r(n)$) is the left (resp.\ right) descendant of $n$ and $\sets A(n)$ is the set of all ancestors of node $n \in \sets B \cup \sets T$. The objective~\eqref{eq:flow_reg_all_obj} maximizes the number of correctly classified datapoints. Note that we can add a regularization term in the objective to control for overfitting, see \citet{aghaei2021strong}. Constraints~\eqref{eq:flow_reg_all_branch_or_predict} ensure that each node is either a branching node, a prediction node, or neither of the two because one of its ancestors is already a prediction node; the last option means that the node is pruned out. Constraints~\eqref{eq:flow_reg_all_terminal_leaf} similarly ensure that each terminal node either makes a prediction or has an ancestor that is a prediction node. Constraints~\eqref{eq:flow_reg_all_conservation_internal} and ~\eqref{eq:flow_reg_all_conservation_terminal} enable flow conservation, whereby every datapoint that flows into node~$n$ must exit to its left (or right) descendant, or flow directly to a sink~$t_k$ that corresponds to a class~$k$. Constraints~\eqref{eq:flow_reg_all_source} ensure that the flow value of each datapoint entering source $s$ is at most 1. Constraints~\eqref{eq:flow_reg_all_branch_left} and \eqref{eq:flow_reg_all_branch_right} enforce datapoints to flow to a node's left (resp. right) child if their branching feature is 0 (resp. 1). Constraints~\eqref{eq:flow_reg_all_sink} require that a datapoint must be directed to the sink corresponding to the class that the prediction node assigns. Finally, constraints~\eqref{eq:flow_reg_all_leaf_prediction} in conjunction with \eqref{eq:integrality_w} ensure that if we make a prediction at node $n$, then exactly one class is associated with its prediction. In sum, each datapoint flows into the graph and lands on exactly one of the sinks via a path from source to sink depending on its feature vector.

\subsection{Fairness Constraints}\label{section:fairness_constraints}

One main advantage of formulation \eqref{eq:flow_reg_all} is its flexible modeling power. In the following, we showcase the variety of constraints that can be added to the formulation to learn trees that satisfy the definitions of fairness we introduced in Section~\ref{section:fairness_def}. We previously defined the various notions of fairness using strict equalities; in practice, however, we may relax this condition by introducing a bias $\delta$, where $\delta$ is the maximum disparity allowed between groups.

\textbf{Statistical Parity.} Statistical parity is satisfied up to a bias of $\delta$ when we add the following constraint to~\eqref{eq:flow_reg_all}:
\begin{equation}\label{eq:statistical_parity}
\begin{split}
     \scriptsize \left|\frac{ \sum_{n \in \sets B \cup \sets T}\sum_{i \in  \sets I: p^i=p} z^i_{n,t_1}}{|\{i \in \sets I : p^i=p\}|} - 
     \frac{ \sum_{n \in \sets B \cup \sets T}\sum_{i \in  \sets I: p^i=p'} z^i_{n,t_1}}{|\{i \in \sets I : p^i=p'\}|}\right| \leq \delta \\
      \forall p,p' \in \sets P: p\neq p',
\end{split}
\end{equation}
where the left-hand side of the inequality is the absolute difference between the proportion of positive classes assigned by the classification tree to groups $p$ and $p'$, respectively. 

\textbf{Conditional Statistical Parity.} In order to describe the CSP constraint, we will let $\ell^i$ denote the value of datapoint $i$'s legitimate factor(s). To ensure that the learned tree satisfies conditional statistical parity up to a bias $\delta$ and for all $\ell \in \sets{L}$, we may augment~\eqref{eq:flow_reg_all} with the constraint
\begin{equation}\label{eq:cond_statistical_parity}
\begin{split}
     \scriptsize \left|\frac{ \sum_{n \in \sets B \cup \sets T}\sum_{i \in  \sets I} \mathbb I(p^i = p \land \ell^i=\ell) z^i_{n,t_1}}{|\{i \in \sets I : p^i=p \land \ell^i=\ell\}|} - \right. \\ \left.
     \frac{ \sum_{n \in \sets B \cup \sets T}\sum_{i \in  \sets I}\mathbb I(p^i = p' \land \ell^i=\ell) z^i_{n,t_1}}{|\{i \in \sets I : p^i=p' \land \ell^i=\ell \}|} \right| \leq \delta \\ \forall p,p' \in \sets P: p\neq p',\; \ell \in \sets L,
\end{split}
\end{equation}
where the left-hand side of the inequality is the absolute difference between the proportions of positive classes assigned by the classification tree to groups $p$ and $p'$\, respectively, in the training data, and whose legitimate feature(s) equal(s) $\ell$.

\textbf{Equalized Odds.} We may augment~\eqref{eq:flow_reg_all} to satisfy equalized odds up to a bias $\delta$ using the constraint
\begin{equation}\label{eq:equalized_odds}
\begin{split}
     \displaystyle \left|\frac{ \sum_{n \in \sets B \cup \sets T}\sum_{i \in  \sets I} \mathbb I(p^i = p \land y^i=k) z^i_{n,t_1}}{|\{i \in \sets I : p^i=p \land y^i=k\}|} - \right. \\ \left.
     \frac{ \sum_{n \in \sets B \cup \sets T}\sum_{i \in  \sets I}\mathbb I(p^i = p' \land y^i=k) z^i_{n,t_1}}{|\{i \in \sets I : p^i=p' \land y^i=k \}|}\right| \leq \delta \\ \forall k \in \sets K, p,p' \in \sets P: p\neq p',
\end{split}
\end{equation}

where the left-hand side of the inequality is the absolute difference between the proportions of false positive and true positive assignments made by the classification tree in groups $p$ and $p'$. Note that predictive equality and equal opportunity are equivalent to setting $k=0$ and $k=1$ in~\eqref{eq:equalized_odds}, respectively.

Note that any of the above constraints can be easily linearized by decomposing them into two. In the general case where we have constraint $|f(x)| \leq \delta$, we can reformulate it into $f(x) \leq \delta$ and $-f(x) \leq \delta$. If $f$ is affine (as is the case here), then these constraints are linear in the decision variables of the problem and the resulting problem can be solved with powerful off-the-shelf mixed-integer \emph{linear} optimization solvers such as Gurobi\footnote{See \url{https://www.gurobi.com/products/gurobi-optimizer/}}. We also emphasize that the above constraints are illustrative examples of the approach's flexibility, and that they may be amended or combined with other fairness considerations. For example, instead of restricting the absolute value of the difference between groups, we may impose that the minority group should be better off by a margin $\delta$, see Section~\ref{section:methodology}.

\section{Experiments}\label{section:experiments}

We now evaluate the empirical performance of the model outlined in Section~\ref{section:formulation}. We first compare the interpretability of several popular machine learning models based on three desiderata -- decision complexity, simulatability, and scope -- illustrating that our approach (\textit{FairOCT}) yields one of the most interpretable models. Then, we compare our approach to a suite of methods for learning fair classifiers (both ones that are model agnostic and those that specifically learn decision trees).

\subsection{Datasets}\label{section:datasets}

\textit{COMPAS.} The Correctional Offender Management Profiling for Alternative Sanctions (COMPAS) is a popular dataset used originally to predict a criminal's risk of recividism after 2 years. \citet{angwin_larson_mattu_kirchner_2016} published a seminal article analyzing that the algorithm deployed then was biased in favor of White criminals. We therefore let race be the sensitive attribute. The original race attribute has 6 levels, but Asian and Native American criminals are quite rare in the data, so we group them under the ``Other'' category in order to make it possible to obtain better estimates of fairness metrics. The dataset consists of 6,172 datapoints.

\textit{Adult.} The UCI Adult dataset is taken from a 1994 Census database \cite{Dua:2019}. The goal is to predict whether or not someone's income exceeds \$50,000 per year, and we treat sex as the sensitive attribute in accordance with \cite{agarwal2018reductions, grari2019fair, kamishima2012fairness} with females being the marginalized group. The full dataset contains 30,162 datapoints.

\textit{German.} The German dataset classifies people as having good or bad credit. We use age as the sensitive attribute, following the works of \cite{kamiran2010discrimination, feldman2015certifying} to split the population into people 25 and younger and older than 25, with the former as the marginalized group since younger people are often assigned worse credit under the basis of age. The dataset contains 1,000 datapoints.

All three datasets are popular benchmark datasets in the algorithmic fairness literature. We focus on these three because they each have different and societally important prediction tasks.

\subsection{Benchmark Methods}\label{section:benchmark_methods}

\textbf{Fair Decision Tree-Based Methods} 

\textit{Optimal and Fair Decision Trees via Regularization (RegOCT)}. We compare our approach to the method for learning optimal fair trees proposed by \citet{aghaei2019learning}, which considers both disparate impact and disparate treatment. In our setting, we are only interested in disparate impact, which reduces to statistical parity in the case of binary outcomes. Our method differs from \textit{RegOCT} in two ways: \textit{1)} \textit{FairOCT} is formulated much more efficiently, resulting in faster solve times, and \textit{2)} \textit{RegOCT} promotes fairness via a regularization term in the objective rather than as a constraint (like ours). Its objective function minimizes misclassification rate plus a regularization term controlled by a ``fairness parameter'' $\rho$. 


\textit{Discrimination Aware Decision Trees (DADT).}
The second tree-based method we compare to is \textit{DADT} proposed by \citet{kamiran2010discrimination}, which consists of a two-pronged approach. The first is an in-processing step that uses sensitivity gain (IGS) and information gain (IGC) as heuristics to build a fair decision tree. If the tree still has a discrimination level greater than $\epsilon$, a post-processing step can be used that relabels the predictions of the tree's leaf nodes until the discrimination reaches $\epsilon$. Note that the original paper proposes to use a combination of these steps where appropriate, such as IGC+IGS, IGC-IGS, or only IGC (which is equivalent to the CART algorithm) -- all of which can be combined with the relabeling step post-training (``Relab''). We compare our method with respect to IGC+IGS\_Relab because \citet{kamiran2010discrimination} cited that it in general performs best, although we display results for all three heuristics in the Appendix Section~\ref{appsec:compare_dadt}. We grow trees of maximum depth 2 and 3 for interpretability. \textit{DADT} has two modeling limitations: it only considers statistical parity as its fairness metric, and it assumes only two sensitive groups exist. In further contrast to \textit{FairOCT}, \textit{DADT} subtracts the probability of the marginalized group from the dominant group. A ``fair'' result under this definition may include a scenario in which the marginalized group receives better outcomes at a much higher rate than the dominant group, which constraint~\eqref{eq:statistical_parity} does not allow. 
However, since our approach is very flexible, we run experiments on \textit{FairOCT} with this notion of disparity -- refer to Appendix Section~\ref{appsec:compare_dadt}.

Note that methods other than \textit{DADT} and \textit{RegOCT} in the literature either: consider individual fairness as opposed to group fairness (e.g., \citet{ranzato2021fair}); learn trees in an online setting (e.g., \citet{zhang2019faht}); or are not interpretable (e.g.,~\citet{grari2019fair}).

\noindent \textbf{Model Agnostic Fairness Methods}

Apart from strictly tree-based methods, we also compare our approach to three methods -- one from each umbrella of approaches: pre-processing, in-processing, and post-processing. Critically, these three methods can learn most ML model classes (such as logistic regression, random forests, kNN, etc.). This feature is important because our goal is to analyze the accuracy and discrimination trade-offs when opting to use our method (one of the most interpretable) in lieu of more complex models.

\textit{[Pre-Processing] Correlation Remover (CR).} We benchmark our method against \textit{CR}, which reduces the correlation between the data and the sensitive feature by regressing the ``centered'' sensitive feature with the non-sensitive features. A linear transformation using the learned coefficients is then applied, resulting in new data $X_{\rm{corr}}$. The final transformation $X_{\rm{tfm}}$ is controlled via a fairness parameter $\alpha \in [0, 1]$:

$$X_{\rm{tfm}} = \alpha X_{\rm{corr}} + (1-\alpha) X,$$

where $\alpha = 0$ corresponds to the original feature vector $X$ and $\alpha = 1$ means all correlation is removed. Further detail can be found in \cite{bird2020fairlearn} and the \href{https://fairlearn.org/main/user_guide/mitigation/preprocessing.html}{API documentation for FairLearn}. Note that we chose this method out of all other pre-processing approaches because it is lightweight and yields better performance upon testing.

\textit{[In-Processing] Exponentiated-Gradient Reduction (ExpG).} Algorithm 1 in \citet{agarwal2018reductions} finds a classifier and $\lambda$ with the highest accuracy subject to a fairness constraint, where $\lambda$ is a vector consisting of $k$ Lagrangian multipliers, each corresponding to a fairness constraint in the algorithm (i.e., $\lambda \in \mathbb{R}_{+}^k$). Since we are interested in obtaining a breadth of accuracy-discrimination datapoints, we opted to implement their ``grid search'' method, which searches through a grid of $\lambda$ and yields the best estimator from a given model class. In our case, we use \textit{all} the results from the possible values of $\lambda$. This grid search method is what we will refer to as \textit{ExpG} in our experiments. We chose to compare our method with \textit{ExpG} because it is one of the only in-processing approaches that take in nearly any model class and also optimize for many fairness notions.

\textit{[Post-Processing] Randomized Threshold Optimizer (RTO)}. Lastly, we compare \textit{FairOCT} with a method proposed by \citet{hardt2016equality} that applies a randomized thresholding transformation to the classifier's prediction to enforce a fairness notion. We refer to Section 3.2 of~\citet{hardt2016equality} for a full treatment of the algorithm. Similar to \textit{ExpG}, we chose \textit{RTO} because it is one of the only post-processing methods that is model agnostic and can optimize for all the fairness notions we consider in this work.
\begin{table*}[t]
\centering
\begin{tabular}{lp{3cm}p{3cm}p{3cm}ll}
\toprule
                \textbf{Model} & Decision Complexity (COMPAS) & Decision Complexity (Adult) & Decision Complexity (German) & Simulatability &  Scope \\
\midrule
        \textbf{Full Tree (d=1)} &                 3 &                3 &                 3 &           High & Global \\
        \textbf{Full Tree (d=2)} &                 7 &                7 &                 7 &           High & Global \\
        \textbf{Full Tree (d=3)} &                15 &               15 &                15 &           High & Global \\
  \textbf{Logistic Regression} &      6 &     12 &      20 &           Medium-High & Global \\
  \textbf{Decision Tree} &                 779 &                5,101 &                 329 &           Medium & Global \\
                  \textbf{k-Nearest Neighbors} &                 4,629 &                22,621 &                 750 &         Medium-Low &  Local \\
     \textbf{SVM (RBF Kernel)} & 3,255 &  10,542 &   454 &            Low &  Local \\
     \textbf{Multilayer Perceptron} &               700 &              1,300 &               2,100 &            Low &   None \\
        \textbf{Random Forest} &                 59,126 &                273,861 &                 17,956 &            Low &  None \\
\bottomrule
\end{tabular}
\caption{Interpretability of various machine learning models with respect to three desiderata: decision complexity (in the context of the \textit{COMPAS}, \textit{Adult}, and \textit{German} datasets), simulaltability, and scope.}
    \label{tab:interpretability_models}
\end{table*}
\subsection{Interpretability of Machine Learning Models}
Table~\ref{tab:interpretability_models} provides a comparison of several popular machine learning models and our assessment of their interpretability with respect to several desiderata: model complexity (defined in Section~\ref{sec:decision_complexity}), simulatability, and scope. In the following, we elucidate our judgments on the simulatability and scope of these models. We will classify models as having ``High'' to ``Low'' simulatability, and ``Global'', ``Local'', or ``None'' in terms of scope. Unless otherwise noted, all models except for optimal trees are trained using the standard parameters of the \href{https://scikit-learn.org/stable/}{scikit-learn package}:

\begin{itemize}
    \item \textbf{Full Binary Trees}: For simplicity, we consider learning full binary trees up to a fixed depth $d$ for our optimal tree methods, although in practice we may easily prune the tree via regularization. We argue that trees have a global scope and are among the most simulatable because humans can visualize the entire decision rule and trace predictions with relative ease (especially when the trees are shallow and simple).
    \item \textbf{Decision Trees -- CART (DT)}: Trees that are grown via a heuristic typically have a stopping point to avoid overfitting, and in our case we impose the minimum number of samples in each leaf to be 3. Since there is stochasticity in tree depth and pruning, we report decision complexity as the average over all trees grown for a particular dataset. While DT has a global scope similar to full binary trees, it is much less simulatable given how deep the trees tend to grow.
    \item \textbf{Logistic Regression (LR)}: Assuming full linearity, logistic regression has global scope because one can wholly observe its decision rule via its coefficients. However, we argue that logistic regression has medium simulatability because the coefficients need to be calculated and contextualized for full interpretability.
    \item \textbf{k-Nearest Neighbors (kNN)}: While kNN's simulatability can be high -- especially in lower dimensions where the decision space can be visualized ($F \leq 3$) -- kNN's are often not simulatable in practice. Without visualizing the entire feature space, we can only observe the local behavior of the model, i.e., which $k$ points are nearest to our new datapoint.
    \item \textbf{Support Vector Machines (SVM), RBF Kernel}: Due to stochasticity, we report the average number of support vectors in our experiments for SVM's decision complexity. We argue that SVM's have low simulatability, especially when $F > 2$ and the decision space is hard to visualize. We can, however, classify SVM's as having local scope because the intuition is similar to kNN -- support vectors closer to the new datapoint will have much higher weights.
    \item \textbf{Random Forest (RF)}: We grow 100 trees via the CART algorithm to build the random forest model, and argue that simulatability is low because a human cannot keep track of the complicated decisions that are aggregated over numerous trees. The intricacy of the random forest also means that it is not interpretable neither globally or locally.
    \item \textbf{Multilayer Perceptron (MLP)}: A multilayer perceptron is a neural network. We train MLPs with 1 hidden layer consisting of 100 nodes, while the input layer has the number of nodes equal to the number of features in the dataset. Its decision complexity, therefore, is equal to $100 \times F + 100$. Since the prediction relies on potentially thousands of matrix calculations, MLPs have very low simulatability and have neither global nor local interpretability.
\end{itemize}

\subsection{Methodology}\label{section:methodology}
\begin{table*}[ht]
\centering
\begin{tabular}{lP{2.3cm}P{1.5cm}P{1.9cm}P{2.2cm}P{2.2cm}P{1.0cm}}
\toprule
                \textbf{Algorithm} & Fairness Definitions & \# Sensitive Levels & Disparity & Model &  Fairness Parameter & Time Limit \\
\toprule
        \textbf{FairOCT} &   SP, CSP, PE, EOpp, EOdds   &      Any  &    absmax &           Full Tree ($d \in \{1, 2, 3\}$) & $\delta \in [0.01, 0.55]$ $\Delta = 0.01$ & 3 hrs \\
\midrule
        \textbf{RegOCT} &   SP  &                Any &                 absmax &           Full Tree ($d \in \{1, 2, 3\}$) & $\rho \in [0, 1]$ $\Delta = 0.02$ & 3 hrs \\
\midrule
        \textbf{DADT} &    SP &               2 &    dom-marg &           DT & $\epsilon \in [0.01, 0.55]$ $\Delta = 0.01$ & N/A \\
\midrule
        \textbf{CR} &  SP, PE, EOpp, EOdds &                Any &                 absmax &           DT, LR, kNN, SVM, RF, MLP & $\alpha \in [0, 1]$ $\quad \quad \Delta = 0.1$ & N/A \\
\midrule
  \textbf{ExpG} &   SP, PE, EOpp, EOdds &     Any &      absmax &           DT, LR, kNN, SVM, RF, MLP & \# of $\lambda: 100$ & N/A \\
  \midrule
     \textbf{RTO} & SP, PE, EOpp, EOdds &  Any &   absmax &            DT, LR, kNN, SVM, RF, MLP &  N/A & N/A \\
\bottomrule
\end{tabular}
\caption{List of algorithms run on \textit{COMPAS}, \textit{Adult}, and \textit{German} datasets. Each algorithm can accommodate different fairness definitions: statistical parity (SP), conditional statistical parity (CSP), predictive equality (PE), equal opportunity (EOpp), and equalized odds (EOdds). They also take varying number of sensitive levels and disparity measures, where ``absmax'' corresponds to the absolute value of the maximum pairwise disparity between sensitive groups and ``dom-marg'' is the difference in positive classification rate between the dominant group and the marginalized group. Different machine learning models are also considered: full binary trees with branching depths $d \in \{1, 2, 3\}$, decision trees grown via CART (DT), logistic regression (LR), k-nearest neighbors (kNN), support vector machine with RBF kernel (SVM), random forest (RF), and multilayer perceptron (MLP). Our experiments also varied the fairness parameters for all methods, whose role is outlined in detail in Section~\ref{section:benchmark_methods}. $\Delta$ denotes the increments taken in each of these fairness parameters. }
    \label{tab:experiments}
\end{table*}

We compare the performance of \textit{FairOCT} with the benchmark methods outlined in Section~\ref{section:benchmark_methods} on all three datasets: \textit{COMPAS}, \textit{German} and \textit{Adult}. Where relevant and possible, we incorporate various fairness definitions in all our experiments. All datasets are tested on statistical parity, predictive equality, equal opportunity, and equalized odds, and \textit{COMPAS} additionally considers conditional statistical parity (conditioned on the number of prior crimes). However, note that not all methods can accommodate all these notions. For instance, only our method can incorporate conditional statistical parity. On the other hand, \textit{RegOCT} and \textit{DADT} can only consider statistical parity. \textit{DADT} additionally only takes in two sensitive levels while all the other methods are robust to more than two sensitive levels. Therefore, in our experiments on \textit{COMPAS}, \textit{DADT} only considers binary race -- ``Black'' vs.\ ``non-Black'' -- which yields results that are not fully comparable to other methods. 

A full summary of our experiments for each method, including capabilities, ML models trained, and fairness parameters tested is provided in Table~\ref{tab:experiments}. We consider fairness parameters in increments of $\Delta$ and display results for all these values in order to fully observe the accuracy-discrimination trade-off. We use 75\% of the data for training and 25\% for testing on all three datasets outlined in Section~\ref{section:datasets} -- split five times with different seeds to account for random variability -- and evaluate model performance on the test set (i.e., out-of-sample or OOS set). The \textit{Adult} dataset is both large and high-dimensional, so we train \textit{FlowOCT} and \textit{RegOCT} on 2,700 datapoints due to computational limitations (while evaluating on the same test set as other methods to allow for an equivalent comparison); we will further discuss this choice in Section~\ref{section:results}. All experiments have a time limit of 3 hours, and utilized four Intel Xeon E5-2640 v4, 2.40GHz CPUs, each having 4GB of memory.

\subsection{Results}\label{section:results}

\textbf{Comparison of Fairness Methods.}
\begin{figure*}
\begin{center}
\centerline{\includegraphics[width=0.85\textwidth]{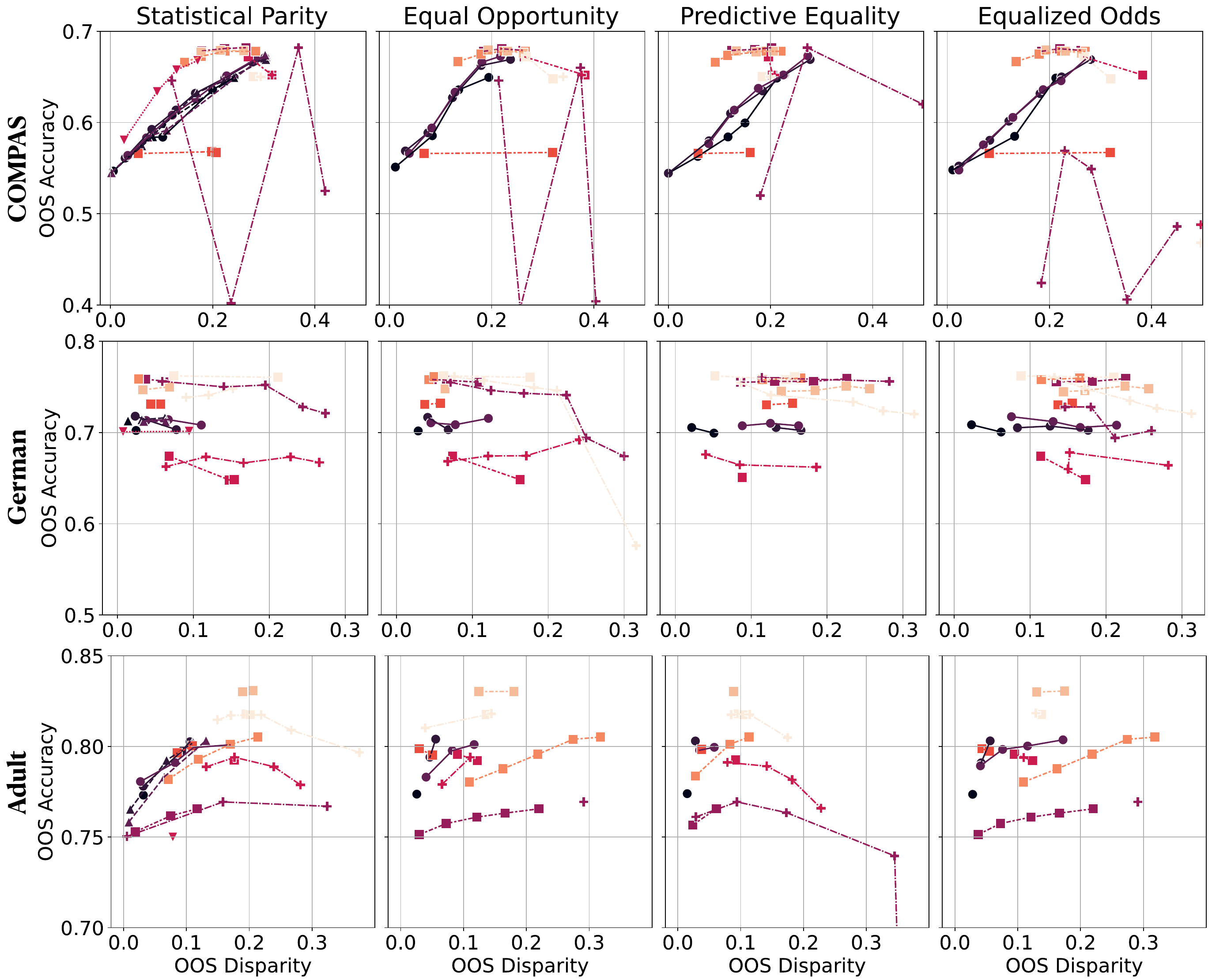}
}
\centerline{\includegraphics[width=0.26\textwidth]{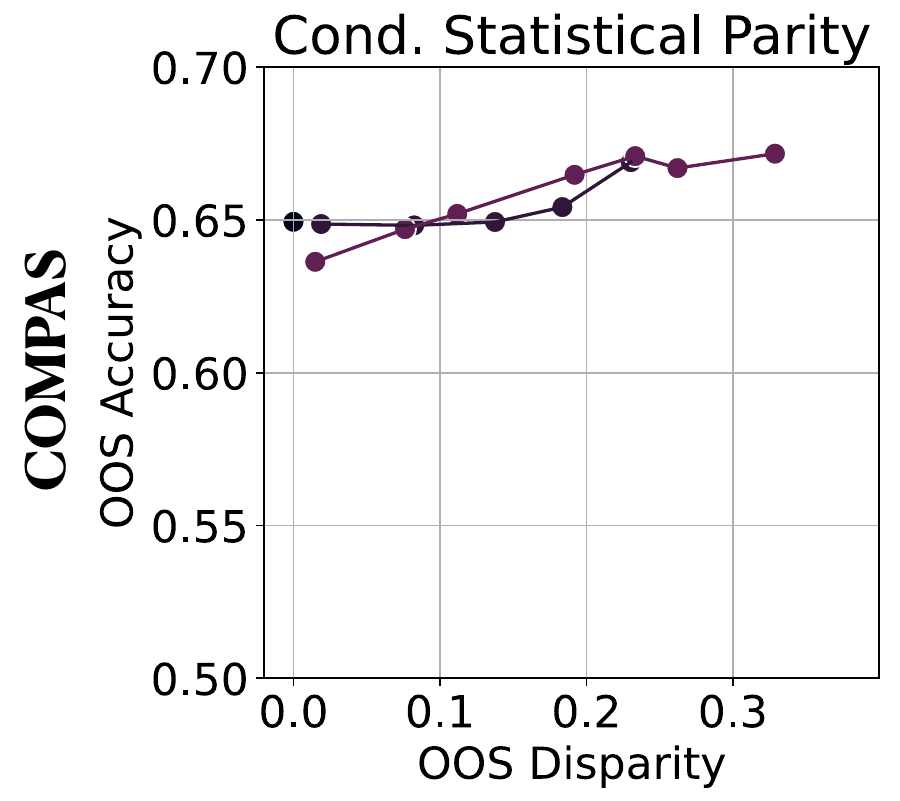}
}
\centerline{\includegraphics[width=0.85\textwidth]{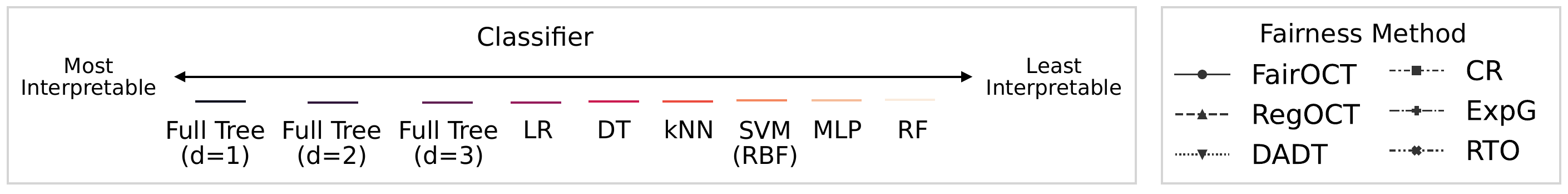}
}
\caption{Accuracy and discrimination of \textit{FairOCT} on the \textit{COMPAS} (top), \textit{Adult} (middle), and \textit{German} (bottom) datasets with varying fairness metrics (from left to right -- statistical parity equal opportunity, predictive equality, and equalized odds, and at the bottom, conditional statistical parity), averaged for each fairness parameter over 5 random samples. Each datapoint in the graph corresponds to an average over 0.05 increments of disparity (i.e., for every method and model, each datapoint is averaged from [0, 0.05), [0.05, 0.1), and so on). Classifiers are ordered from darkest (most interpretable) to lightest (least interpretable) according to our interpretability desiderata outlined in Table~\ref{tab:interpretability_models}. Different fairness methods are outlined by different marker and line styles. For ease of visualization, we average the results for \textit{DADT} with trees of maximum depth 2 and 3.}
\label{fig:experiments}
\end{center}
\end{figure*}
Figure~\ref{fig:experiments} plots the accuracy and discrimination trade-off for all experiments outlined in Table~\ref{tab:experiments}. In general, \textit{FairOCT} trained on all depths ($d=\{1, 2, 3\}$) obtain similar performances, though this is mainly attributed to the experiments on $d=3$ not solving to optimality within the time limit. Trends vary across datasets; for instance, \textit{FairOCT} found decisions with accuracies between $[0.55, 0.68]$ with disparities between $[0, 0.25]$ for \textit{COMPAS}. In contrast, its results for \textit{Adult} varied only by at most 5 percentage points (p.p.) in accuracy and 10 p.p. in disparity. Nonetheless, \textit{FairOCT} learns a range of heterogeneous performances, which allows us to choose between various accuracy-discrimination trade-offs.

Overall, for any given disparity threshold, \textit{FairOCT} consistently outperforms \textit{DADT} within the notion of statistical parity (recall that \textit{DADT} is only trained on this notion). This is expected since \textit{FairOCT} finds an optimal solution whereas \textit{DADT} relies on a heuristic. A notable aberration is \textit{DADT}'s superior performance to \textit{FairOCT} on \textit{COMPAS}, but this is because \textit{DADT} only considers two levels of race while all other experiments have four levels. Even when only considering two levels of race, \textit{DADT} does not result in significant improvements in performance. Refer to Appendix Section~\ref{appsec:compare_dadt} for a comparison of \textit{FairOCT} and \textit{DADT} within the same learning setting, which shows that \textit{FairOCT} consistently outperforms \textit{DADT}.

In contrast, the other tree-based benchmark method -- \textit{RegOCT} -- performs at around the same level as \textit{FairOCT} (again only on statistical parity because \textit{RegOCT} only considers this notion). This result is unsurprising given that \textit{RegOCT} similarly learns optimal trees, so its performance will be as good as \textit{FairOCT}, which already finds the best-performing partition for a fixed maximum depth. However, we emphasize that \textit{RegOCT} is much less flexible in considering other notions of fairness. Moreover, \textit{FairOCT} benefits from a stronger formulation and is faster by at least an order-of-magnitude. Refer to Appendix Table~\ref{tab:comp_time} for a full comparison of computational times.

With regard to non-tree-based methods, \textit{CR} in general yields smooth and heterogeneous results throughout the disparity axis. We observe the same trend for \textit{ExpG}, with the exception of the \textit{COMPAS} dataset, whose performance jumps widely. \cite{agarwal2018reductions} mentioned that as the number of sensitive levels (and therefore the number of constraints) increase, \textit{ExpG} is not recommended because the search space grows exponentially. This limitation may be circumvented by testing an arbitrarily large number of fairness parameters and ignoring several outliers to smooth the curve. Note also that most of the parameters tested led to a disparity level of 1.0 or near 1.0, but we decided to truncate the results to a more reasonable level for all datasets. Finally, \textit{RTO} is mainly missing from Figure~\ref{fig:experiments} because we cannot tune any fairness parameters -- resulting in a single datapoint, none of which is displayed because the performance hovers at around 0.6-0.9 disparity.

In comparison, \textit{FairOCT} consistently finds results in the lower end of the disparity axis (which is the space we care most given the task of promoting full parity). For instance, in the \textit{COMPAS} dataset on all the fairness notions, \textit{FairOCT} resulted in disparities ranging from 0 to 0.3 -- no other method boasts this range. Even in the \textit{Adult} and \textit{German} datasets where that range is attenuated, no other fairness-promoting method consistently finds datapoints with low disparity (i.e., $< 0.05$). This result likely arises from our method taking in a hyperparemeter $\delta$ that upper bounds resulting disparity. The optimization problem is then forced to find the highest-accuracy decision rule given a (potentially low) tolerance on disparity. This is in contrast to all other methods; RegOCT uses a regularization term, CR tunes the extent of correlation removed, and ExpG passes numerous values of $\lambda$, all of which do not guarantee certain disparity results. 

Finally, we note that the \textit{FairOCT} results trained on the 2,700 datapoints of \textit{Adult} have comparable performances with other baseline methods, indicating that while optimization-based methods like \textit{FairOCT} may run into computational problems, training on a smaller, representative dataset yields predictions that generalize very well to the entire population. In the bottom graph of Figure~\ref{fig:experiments}, we also conduct experiments where \textit{FairOCT} considers conditional statistical parity on \textit{COMPAS}; no other method we compare to can equivalently compare this fairness notion, highlighting our approach's flexible modeling power. 

\noindent \textbf{The Price of Interpretability.}
As expected, more interpretable models like full trees perform (marginally) worse than more complex models like random forests and MLP's. This difference denotes the price of interpretability. For instance, in the \textit{German} dataset, a random forest trained on \textit{CR} has on average 6 percentage points higher accuracy than a full tree of depth 2 trained on \textit{FairOCT}, across all discrimination thresholds. However, as mentioned previously, many fairness-promoting methods fail to reach full or near perfect parity in contrast to FairOCT, rendering this aggregated comparison less meaningful. Notably, in the \textit{COMPAS} dataset, a neural net trained on CR might perform on average 3 percentage points better than a \textit{FairOCT} ($d=2$), but only within the disparity ranges for which both methods have results. We must further consider the other datapoints \textit{FairOCT} ($d=2$) yields for disparities less than 0.15, of which have no equivalent comparison with the neural net. Nonetheless, given a fixed disparity threshold, the best performing complex model performs on average 4.2 p.p. better in terms of OOS accuracy than \textit{FairOCT} over the range of disparities for which both models have results.

\section{Conclusion}
In this work, we presented an MIO formulation for learning optimal classification trees that can be modeled to consider a variety of algorithmic fairness notions. We also propose a new measure of interpretability named \textit{decision complexity} in order to compare our interpretable method with other classes of models. In doing so, we conduct one of the first experiments that analyze in-depth the trade-offs between interpretability, fairness, and predictive accuracy.

Our experiments show that while we observe a (often small) price of interpretability with trees of shallow depth, one must ultimately consider not only the trade-offs between interpretability, accuracy, and discrimination, but also the various fairness-promoting methods that may yield vastly different results (e.g., with regard to the range of disparities found, how different methods might be better for certain constraints, etc.). In reality, decision makers face the incredibly hard task of balancing these trade-offs given the application at hand. Our work attempts to enrich these considerations in the hopes of guiding practitioners when they make these difficult decisions.

\begin{acks}
N.\ Jo acknowledges support from the Epstein Institute at the University of Southern California. P.\ Vayanos and S.\ Aghaei are funded in part by the National Science Foundation under CAREER award number 2046230. They are grateful for this support. N.\ Jo, P.\ Vayanos, and S.\ Aghaei gratefully acknowledge support from the Hilton C.\ Foundation, the Homeless Policy Research Institute, and the Home for Good foundation under the ``C.E.S.\ Triage Tool Research \& Refinement'' grant. A.\ G\'omez is funded in part by the National Science Foundation under grant 2006762.
\end{acks}

\bibliographystyle{ACM-Reference-Format}
\bibliography{bib}

\clearpage
\appendix

\begin{figure*}[b]
\begin{center}
\includegraphics[width=0.8\textwidth]{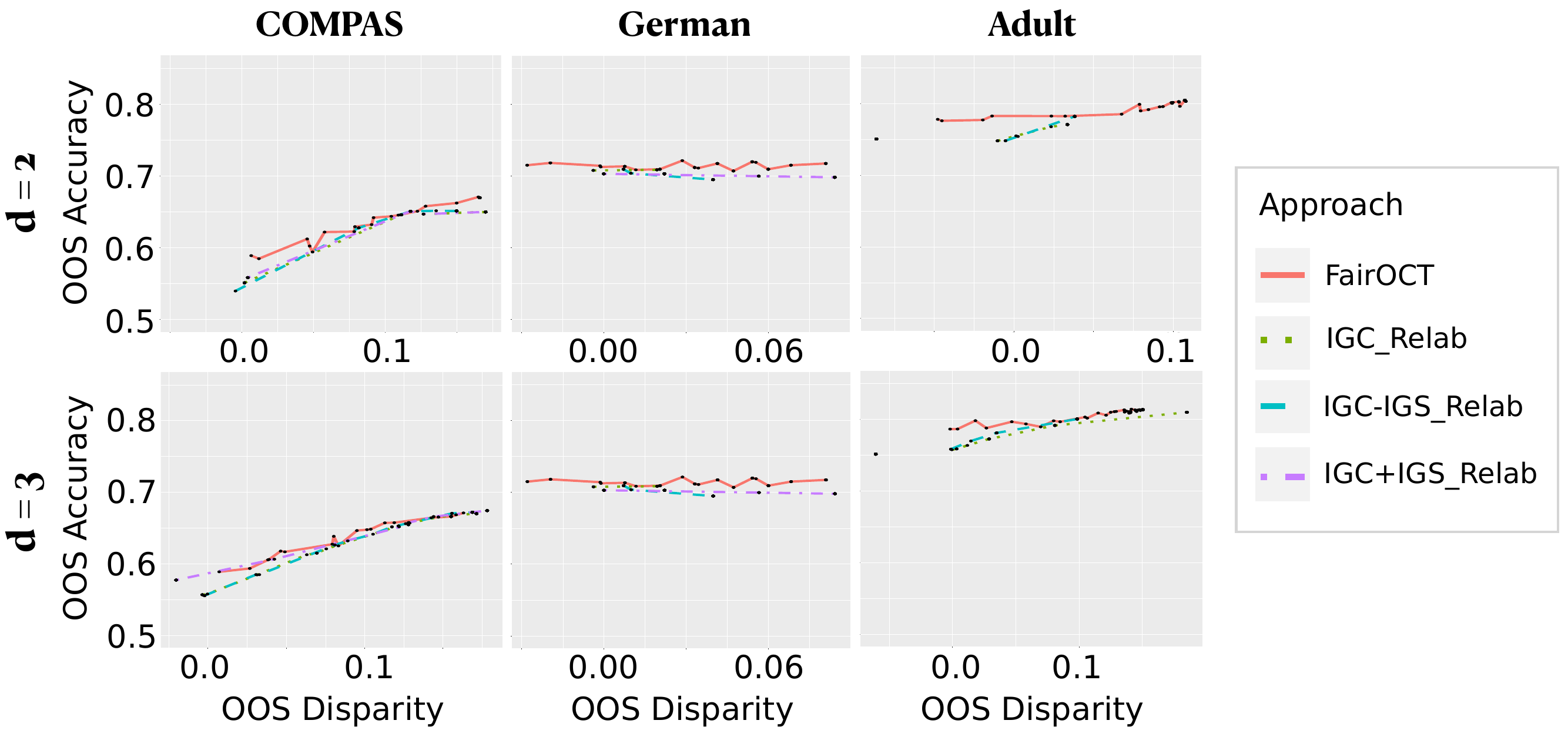}
\caption{Comparison of the accuracy and discrimination of \textit{FairOCT} and \textit{DADT} for trees of depth 2 (top) and 3 (bottom) when the discrimination bounds ($\delta$) are varied on three datasets: \textit{COMPAS} (left column), \textit{German} (middle column), and \textit{Adult} (right column) -- averaged over 5 random train-test splits. }
\label{fig:compare_to_kamiran}
\end{center}
\end{figure*}

\section{Equivalent Comparisons to DADT}\label{appsec:compare_dadt}

In order to compare \textit{FairOCT} fairly with \textit{DADT}, we collapse the protected variables in our datasets to two groups where applicable and only consider statistical parity as our fairness measure. We also adapt our statistical parity constraint~(\ref{eq:statistical_parity}) to fit the fairness definition as implemented by \textit{DADT}:
\begin{equation*}
\begin{split}
     \displaystyle \frac{\displaystyle \sum_{n \in \sets B \cup \sets T}\sum_{i \in  \sets I: p^i=p_d} z^i_{n,t_1}}{|\{i \in \sets I : p^i=p_d\}|} - 
     \frac{\displaystyle \sum_{n \in \sets B \cup \sets T}\sum_{i \in  \sets I: p^i=p_m} z^i_{n,t_1}}{|\{i \in \sets I : p^i=p_m\}|} \leq \delta,
\end{split}
\end{equation*}
where $p_d$ and $p_m$ refer to the dominant and marginalized groups, respectively. The only dataset that requires further processing is \textit{COMPAS}, which has multiple levels for race. We will let the sensitive groups be White versus non-White criminals, White being the dominant group. For both \textit{DADT} and \textit{FairOCT}, we grow trees of depths $d \in \{2, 3\}$ in order to produce interpretable results.

Figure~\ref{fig:compare_to_kamiran} shows the accuracy and discrimination for both \textit{FairOCT} and \textit{DADT} when we vary the discrimination threshold $\delta$, on trees of depths 2 and 3. Overall, for any given discrimination level, \textit{FairOCT} consistently has better in-sample and out-of-sample accuracies. This is expected since \textit{FairOCT} finds an optimal solution whereas DADT relies on a heuristic. Given a fixed discrimination threshold, \textit{FairOCT} improves out-of-sample (OOS) accuracy by 1.6, 2.8, 1.7 percentage points on average over IGC\_Relab, IGC+IGS\_Relab, and IGC-IGS\_Relab, respectively. Our method also obtains a higher OOS accuracy in 100\% of the experiments compared to all \textit{DADT} methods.

Since the \textit{Adult} dataset is both large and high-dimensional, \textit{FairOCT} struggled to reach an optimal solution on the full training set within the time limit, which resulted in suboptimal performances throughout. Therefore, the results shown in Figure~\ref{fig:compare_to_kamiran} are from a set of experiments where \textit{FairOCT} learns from 2,700 datapoints but is evaluated on the same test set to allow for a fair comparison with \textit{DADT}. Similar to the main set of experiments, we find that the optimal solution in fact consistently outperforms \textit{DADT}. This means that, in spite of the difficulties in solving MIO problems, opting for the optimal solution (even on a smaller training set) may yield better results than relying on a heuristic.

\section{Computational Times}
Table~\ref{tab:comp_time} outlines the average and standard deviation of computational times over all experiments, on all methods and fairness parameters. Naturally, obtaining provably optimal solutions comes at a computational cost. \textit{FairOCT} has average solve times of 1.1 hours for \textit{COMPAS}, 1.5 hours for \textit{German}, and 2.5 hours for \textit{Adult}. Notably, however, \textit{FairOCT} is much faster than the other optimization-based method \textit{RegOCT}, which consistently does not solve to optimality for depths 2 and 3 over all datasets. On the other hand, \textit{DADT} and \textit{CR} consistently finds a tree in under 1 second for \textit{COMPAS} and \textit{German}, and around 5 seconds for \textit{Adult}. \textit{RTO} is similarly fast, while \textit{ExpG} is notably slower than these other methods, mainly because of the need to search through many choices of $\lambda$.

\begin{table*}
\centering
\begin{tabular}{lP{3cm}P{3cm}P{3cm}}
\toprule
 & \multicolumn{3}{c}{Mean (std. deviation) of time elapsed (in seconds)} \\
        \textbf{Model} &             COMPAS &               Adult &              German \\
\midrule
\textbf{FairOCT (d=1)} &      144.8 (85.9) &       92.8 (23.9) &        27.3 (5.5) \\
\textbf{FairOCT (d=2)} & 3,014.6 (2,117.3) & 8,371.1 (2,349.4) & 6,644.9 (1,821.0) \\
\textbf{FairOCT (d=3)} &   10,798.0 (90.3) &    10,800.0 (0.0) &    10,800.0 (0.0) \\
\midrule
 \textbf{RegOCT (d=1)} &   2,210.9 (901.5) &   9,515.5 (1,307.3) &   612.4 (200.3) \\
 \textbf{RegOCT (d=2)} &        10,800.0 (0.0) &        10,800.0 (0.0) &        10,800.0 (0.0) \\
 \textbf{RegOCT (d=3)} &        10,800.0 (0.0) &        10,800.0 (0.0) &        10,800.0 (0.0) \\
 \midrule
         \textbf{DADT} &         0.5 (0.3) &         4.5 (2.4) &         0.2 (0.2) \\
           \textbf{CR} &         0.4 (0.4) &         5.7 (7.0) &         0.2 (0.3) \\
         \textbf{ExpG} &       18.3 (11.7) &       82.5 (55.9) &         8.0 (5.8) \\
          \textbf{RTO} &         0.7 (0.6) &        8.8 (11.2) &         0.4 (0.6) \\
\bottomrule
\end{tabular}
\caption{Average and standard deviation computational times over all experiments on all methods and fairness parameters tested. The results for \textit{DADT} are averaged over trees of maximum depth 2 and 3. The results for \textit{CR}, \textit{ExpG}, and \textit{RTO} are averaged over all model classes.}
    \label{tab:comp_time}
\end{table*}

\end{document}